\documentclass{article}
\pdfoutput=1
\usepackage[preprint]{neurips_2023}
\usepackage[utf8]{inputenc} %
\usepackage[T1]{fontenc}    %
\usepackage{xurl}
\usepackage{hyperref}            %
\usepackage{breakurl}          %
\usepackage{amsfonts}       %
\usepackage{amsmath}       %
\usepackage{nicefrac}       %
\usepackage{microtype}      %
\usepackage{lipsum}
\usepackage{graphicx}
\usepackage{makecell}
\usepackage{wrapfig}
\usepackage[size=small]{caption}
\usepackage{mathtools}
\usepackage{amssymb}
\usepackage{amsthm}
\usepackage[algoruled,boxed,lined,noend]{algorithm2e}
\usepackage{adjustbox} 
\usepackage{float}
\usepackage{pifont}
\usepackage{subcaption}
\usepackage{fontawesome}
\usepackage{multirow}
\usepackage{rotating}
\usepackage{array}    
\usepackage{lineno}
\usepackage{longtable}
\usepackage[table,dvipsnames]{xcolor}
\usepackage[numbers]{natbib}
\usepackage{textcomp}
\usepackage{ulem}
\usepackage{color}
\usepackage[most]{tcolorbox}
\usepackage{xcolor}

\definecolor{qual-fig-green}{RGB}{0,144,11}
\definecolor{qual-fig-red}{RGB}{238,0,0}
\definecolor{qual-fig-purple}{RGB}{153,51,255}

\definecolor{answergreen}{RGB}{222,255,222}
\definecolor{promptblue}{RGB}{235,245,255}   
\definecolor{promptgreen}{RGB}{235,255,235}  
\definecolor{promptgray}{RGB}{248,248,248}

\title{Capabilities of GPT-5 on Multimodal Medical Reasoning}

\author{
Shansong Wang$^{1}$  \quad Mingzhe Hu$^{1}$ \quad Qiang Li$^{1}$ \quad Mojtaba Safari$^{1}$ \quad   \textbf{Xiaofeng Yang}$\textsuperscript{1 \faEnvelope}$ 
\vspace{3mm}
\\
$^1$Department of Radiation Oncology, Winship Cancer Institute, Emory University School of Medicine\\
\faEnvelope \quad Corresponding author: xiaofeng.yang@emory.edu \\
}

\begin{document}
%\linenumbers
\maketitle

\begin{abstract}
    Recent advances in large language models (LLMs) have enabled general-purpose systems to perform increasingly complex domain-specific reasoning without extensive fine-tuning. In the medical domain, decision-making often requires integrating heterogeneous information sources, including patient narratives, structured data, and medical images. This study positions GPT-5 as a generalist multimodal reasoner for medical decision support and systematically evaluates its zero-shot chain-of-thought reasoning performance on both text-based question answering and visual question answering tasks under a unified protocol. We benchmark GPT-5, GPT-5-mini, GPT-5-nano, and GPT-4o-2024-11-20 against standardized splits of MedQA, MedXpertQA (text and multimodal), MMLU medical subsets, USMLE self-assessment exams, and VQA-RAD. Results show that GPT-5 consistently outperforms all baselines, achieving state-of-the-art accuracy across all QA benchmarks and delivering substantial gains in multimodal reasoning. On MedXpertQA MM, GPT-5 improves reasoning and understanding scores by +29.26\% and +26.18\% over GPT-4o, respectively, and surpasses pre-licensed human experts by +24.23\% in reasoning and +29.40\% in understanding. In contrast, GPT-4o remains below human expert performance in most dimensions. A representative case study demonstrates GPT-5's ability to integrate visual and textual cues into a coherent diagnostic reasoning chain, recommending appropriate high-stakes interventions. Our results show that, on these controlled multimodal reasoning benchmarks, GPT-5 moves from human-comparable to above human-expert performance. This improvement may substantially inform the design of future clinical decision-support systems. We make the code public at the \href{https://github.com/wangshansong1/GPT-5-Evaluation}{GPT-5-Evaluation}.
\end{abstract}

\section{Introduction}
Rapid iteration of general-purpose large language models (LLMs)~\cite{achiam2023gpt,liu2024deepseek,openai2025gpt5} in recent years has promoted a paradigm shift from ``task-specific models'' to ``LLM as core components''. In medical scenarios, real-world problems often span multiple forms of evidence, including medical history text~\cite{azar2025llm,singhal2023large}, structured indicators~\cite{guo2023gpt4graph}, and medical imaging~\cite{wang2025triad,wang2025unifying}. This requires models to not only understand language, but also perform consistent reasoning and decision-making across heterogeneous modalities~\cite{liu2025application}. Enabling LLMs to reliably perform this type of multimodal medical reasoning without relying on extensive domain-specific fine-tuning is becoming a key issue in medical artificial intelligence (AI)~\cite{nori2023capabilities,thirunavukarasu2023large,hu2024advancing}.  

The release of GPT-3.5~\cite{openai2023gpt35} and GPT-4~\cite{achiam2023gpt} marked the beginning of this turning point. They brought general ``prompt-to-use'' capabilities to specialized tasks, significantly shifting the boundaries of research and application~\cite{liu2025application,thirunavukarasu2023large}. Their robust performance in few-shot/zero-shot settings, stronger instruction following, and dialogue interaction make it possible to handle interdisciplinary problems with a unified interface.  For example, since late 2022, general-purpose assistants built upon these models have garnered significant attention for their impressive out-of-the-box performance on professional and academic benchmarks, including graduate entrance exams and subject assessments~\cite{achiam2023gpt}, even achieving near-passing accuracy on the USMLE without domain-specific fine-tuning~\cite{yang2023performance}. Across clinical specialties (e.g., neurosurgery~\cite{hopkins2023chatgpt}, hepatology~\cite{yeo2023assessing}, and core internal medicine domains~\cite{johnson2023assessing}), they have exhibited promising knowledge recall and reasoning, and early studies have explored decision-support roles in radiology~\cite{bhayana2023gpt}, pathology~\cite{omar2024chatgpt}, and orthodontics~\cite{demir2024enhancing}. In daily clinical workflows, such LLMs can draft clinic letters~\cite{tung2024comparison}, discharge summaries~\cite{kweon2024ehrnoteqa}, and cancer screening plans~\cite{hao2025large}. Yet most prior evaluations remain predominantly text-centric and heterogeneous in datasets, prompting, and scoring, obscuring how these gains translate to settings that require joint reasoning over reports, images, and structured signals. 

To this end, we position GPT-5~\cite{openai2025gpt5} as a generalist multimodal reasoner and evaluate it under a unified protocol to enable controlled, longitudinal comparisons with GPT-4 on accuracy. We further investigate whether a single instruction-following model can serve as a reliable hub for multimodal medical decision support. Concretely, we evaluate GPT-5’s reasoning ability on question answering (QA) and visual question answering (VQA).  We standardize splits and prompts across GPT-4/5, evaluating zero-shot regimes with the same exemplars, chain-of-thought (CoT) supervision, and answers constrained to a single final choice for multiple-choice items. This design isolates the contribution of the model upgrade itself, rather than prompt engineering or dataset idiosyncrasies, in testing whether GPT-5 can act as a reliable hub for multimodal medical decision support.

\section{Methodology}

\subsection{Datasets}
To evaluate GPT-5, we consider four datasets that span both text-based and multimodal medical reasoning tasks. For question answering in text-only settings, we use MedQA\cite{jin2020disease} and the medical subset of MMLU\cite{hendryckstest2021} (MMLU-Medical). For visual question answering, we employ VQA-RAD\cite{lau2018dataset} and the newly introduced MedXpertQA\cite{zuo2025medxpertqa}. Together, these datasets cover a wide range of medical knowledge domains, reasoning types, and input modalities.

\begin{itemize}
    
\item \textbf{MedQA}~\cite{jin2020disease} contains multiple-choice questions in English, Simplified Chinese, and Traditional Chinese, collected from the medical licensing examinations of the U.S., Mainland China, and Taiwan. Each question in the United States and Mainland China subsets has five answer options, with 1,273 and 3,426 questions in the respective test splits. The Taiwan subset contains 1,413 questions with 4 options per question. In addition, a simplified version of the U.S. test split provides 4-option variants of the same questions by removing one incorrect answer choice. Here, we use a simplified version of the US test set for evaluation.

\item \textbf{MMLU}~\cite{hendryckstest2021} is a large-scale multiple-choice benchmark spanning 57 subjects across diverse domains. In this work, we focus on the MMLU-Medical to assess GPT-5’s performance on a broad spectrum of specialized medical knowledge and reasoning skills.

\item \textbf{USMLE Self Assessment}~\cite{kung2023performance} from official practice materials provided by the U.S. Medical Licensing Examination (USMLE) program\footnote{\url{https://www.usmle.org/prepare-your-exam}}. The dataset comprises sample questions from three separate PDF files corresponding to Step 1, Step 2 CK, and Step 3, covering a broad range of clinical knowledge domains. This dataset setup follows the protocol of Harsha et al.~\cite{nori2023capabilities}, who employed these sample exams to evaluate LLM performance on medical licensing assessments. In line with their approach, we preserved the original structure and content of the official sample materials. 

\item \textbf{MedXpertQA}~\cite{zuo2025medxpertqa} is a challenging and comprehensive benchmark designed to evaluate expert-level medical knowledge and advanced reasoning. It comprises 4,460 questions spanning 17 specialties and 11 body systems, with two subsets: a text-only set and a multimodal set. The multimodal subset introduces complex clinical exam questions with diverse medical images, patient records, and examination results, going beyond traditional medical VQA datasets that rely on simplified image-caption pairs. To ensure clinical relevance and difficulty, MedXpertQA incorporates specialty board questions, rigorous filtering, data synthesis to mitigate leakage, and multiple rounds of expert review.

\item \textbf{VQA-RAD}~\cite{lau2018dataset} contains 2,244 question–answer pairs linked to 314 radiology images sourced from the MedPix database. The questions include both open-ended and binary yes/no formats, designed to evaluate visual understanding in clinical radiology contexts. The dataset is widely used for training and testing medical VQA systems and has undergone manual curation by clinicians to ensure quality and clinical validity. Here, we use the binary ``yes/no'' samples in the test set, totaling 251 samples.
\end{itemize}

\subsection{Prompting Design}

We evaluate GPT-5 using a zero-shot CoT approach. In this setting, each interaction is a brief chat that first elicits step-by-step reasoning and then restricts the answer to a discrete choice. A system message anchors the medical domain. The first user turn presents the question and explicitly triggers CoT via ``\textcolor{blue}{Let’s think step by step.}'' The assistant then produces a free-form rationale (stored as \textit{prediction\_rationale}) without committing to an option. A second user turn provides a convergence cue: ``\textcolor{blue}{Therefore, among A through \textcolor{red}{\{END\_LETTER\}}, the answer is}'', where \textcolor{red}{\{END\_LETTER\}} denotes the last option letter computed from the number of choices. The final assistant turn returns the option letter (stored as \textit{prediction}). For multimodal items, all images associated with the sample are appended as \texttt{image\_url} entries to the first user message, enabling the model to reason over text and images within a single turn while keeping the subsequent convergence step purely textual. The JSON templates below instantiate this protocol for the no-image and with-images variants, using \textcolor{red}{\{QUESTION\_TEXT\}}, \textcolor{red}{\{END\_LETTER\}}, \textcolor{red}{\{IMAGE\_URL\_1\}}, \textcolor{red}{\{ASSISTANT\_RATIONALE\}}, and \textcolor{red}{\{ASSISTANT\_FINAL\}} as placeholders. The prompting design template for the QA/VQA task is shown in Fig. \ref{fig:promptVQA}, and a specific example is shown in Fig. \ref{fig:promptVQAsample}.

\begin{figure}[htbp]
\centering
\begin{tcolorbox}[
        colback=promptgray,
        colframe=black!50!black,
        colbacktitle=white!50!black,
        title=Zero-shot + CoT for QA,
        fontupper=\normalsize 
    ]
{\linespread{0.8}\selectfont
\begin{verbatim}
[
  {
    "role": "system",
    "content": "You are a helpful medical assistant."
  },
  {
    "role": "user",
    "content": [
      {
        "type": "text",
        "text": "Q: {QUESTION_TEXT}\nA: Let's think step by step."
      }
    ]
  },
  {
    "role": "assistant",
    "content": "{ASSISTANT_RATIONALE}"
  },
  {
    "role": "user",
    "content": "Therefore, among A through {END_LETTER}, the answer is"
  },
  {
    "role": "assistant",
    "content": "{ASSISTANT_FINAL}"
  }
]
\end{verbatim}
}
\end{tcolorbox}
\begin{tcolorbox}[
        colback=promptgray,
        colframe=black!50!black,
        colbacktitle=white!50!black,
        title=Zero-shot + CoT for VQA,
        fontupper=\normalsize 
    ]
{\linespread{0.8}\selectfont
\begin{verbatim}
[
  {
    "role": "system",
    "content": "You are a helpful medical assistant."
  },
  {
    "role": "user",
    "content": [
      {
        "type": "text",
        "text": "Q: {QUESTION_TEXT}\nA: Let's think step by step."
      },
      {
        "type": "image_url",
        "image_url": { "url": "{IMAGE_URL_1}" }
      }
    ]
  },
  {
    "role": "assistant",
    "content": "{ASSISTANT_RATIONALE}"
  },
  {
    "role": "user",
    "content": "Therefore, among A through {END_LETTER}, the answer is"
  },
  {
    "role": "assistant",
    "content": "{ASSISTANT_FINAL}"
  }
]
\end{verbatim}
}
\end{tcolorbox}

\caption{Prompting design for QA/VQA task}
\label{fig:promptVQA}
\end{figure}

\begin{figure}[htbp]
\centering
\begin{tcolorbox}[
    colback=promptgray,
    colframe=black!50!black,
    colbacktitle=white!50!black,
    title=A Sample from MedXpertQA,
    fontupper=\normalsize,
    enhanced,
    overlay={
        \node[anchor=north east, inner sep=3pt, yshift=-15pt] 
        at (frame.north east) {\includegraphics[width=0.25\textwidth]{MM-1993-a.jpeg}};
    } 
]
{\linespread{0.8}\selectfont
\begin{verbatim}
[
  {
    "role": "system",
    "content": "You are a helpful medical assistant."
  },
  {
    "role": "user",
    "content": [
      {
        "type": "text",
        "text": "
            Q: A 45-year-old man is brought to the emergency department by 
            police after being found unconscious in a store. He is wearing
            soiled clothing that smells of urine, and his pants are soaked
            in vomit. His medical history includes IV drug use, alcohol use
            , and fractures due to scurvy. He is not on any current medica-
            tions. Initial vital signs show a temperature of 99.5°F (37.5°C)
            , blood pressure of 90/63 mmHg, pulse of 130/min, respirations
            of 15/min, and oxygen saturation of 95% on room air. The patient
            is treated with IV fluids, thiamine, and dextrose, after which he
            becomes more alert but continues vomiting. Physical examination 
            reveals epigastric tenderness, while cardiac and pulmonary exams
            are unremarkable. A CT scan of the abdomen is performed, and lab-
            oratory results are as follows:
             - Serum:                      
                Na+: 139 mEq/L                    Creatinine: 1.1 mg/dL            
                Cl-: 102 mEq/L                    Ca2+: 10.2 mg/dL
                K+: 4.0 mEq/L                     Lipase: 295 U/L
                HCO3-: 26 mEq/L                   AST: 57 U/L
                BUN: 20 mg/dL                     ALT: 39 U/L
                Glucose: 73 mg/dL
             - Hematology:                      
                Hemoglobin: 9 g/dL                Hematocrit: 30%
                Mean corpuscular volume: 120 µm³  Leukocyte count: 8,500/mm³
                Platelet count: 199,000/mm³                 
            Several hours later, his vital signs improve to a temperature of 
            99.5°F (37.5°C), blood pressure of 110/72 mmHg, pulse of 97/min, 
            respirations of 15/min, and oxygen saturation of 95% on room air. 
            On examination, suprasternal crepitus is noted, along with blood-
            stained vomitus in the oropharynx. Cardiac and pulmonary findings
            remain normal, and the lower extremities show no abnormalities. 
            What is the most appropriate next step in this patient’s management?            
            Answer Choices: 
                (A) Ondansetron               (B) Folate and vitamin B12 
                (C) Supportive therapy        (D) Injection of epinephrine 
                (E) Gastrografin swallow
            A: Let's think step by step."
      },
      {
        "type": "image_url",
        "image_url": { "url": "images/MM-1993-a.jpeg" }
      }
    ]
  },
  {
    "role": "assistant",
    "content": "{ASSISTANT_RATIONALE}"
  },
  {
    "role": "user",
    "content": "Therefore, among A through E, the answer is"
  },
  {
    "role": "assistant",
    "content": "E"
  }
]
\end{verbatim}
}
\end{tcolorbox}
\caption{A prompting design sample from MedXpertQA.}
\label{fig:promptVQAsample}
\end{figure}

\section{Results}

\subsection{Performance of GPT-5 on QA Benchmarks}

On text-based medical QA datasets (Table \ref{tab:med-text-results}), GPT-5 achieved consistent gains over GPT-4o and smaller GPT-5 variants. On MedQA (US 4-option), GPT-5 reached 95.84\%, a 4.80\% absolute improvement over GPT-4o, indicating stronger factual recall and diagnostic reasoning in clinical question contexts. The most pronounced gains appeared in MedXpertQA Text, where reasoning accuracy improved by 26.33\% and understanding by 25.30\% over GPT-4o. This suggests a substantial enhancement in multi-step inference and nuanced comprehension of medical narratives.
In MMLU medical subdomains, GPT-5 maintained near-ceiling performance ($>$91\% across all subjects), with notable gains in Medical Genetics (+4.00\%) and Clinical Knowledge (+2.64\%). The improvements were generally incremental in high-baseline categories, indicating that GPT-5’s upgrades mainly benefit tasks with higher reasoning complexity rather than purely factual recall.

\begin{table}[htbp]
\centering
\caption{Performance on QA benchmarks (\%). The blue numbers and arrows indicate changes compared to GPT-4o-2024-11-20.}
\label{tab:med-text-results}
\adjustbox{width=\textwidth,center}{
\begin{tabular}{lcccc}
\hline
Dataset & GPT-5 & GPT-5-mini & GPT-5-nano & GPT-4o-2024-11-20 \\
\hline
\multicolumn{5}{l}{\textbf{MedQA}}\\
US (4-option) & \textbf{95.84} (\textcolor{blue}{$\uparrow$4.80\%}) & 93.48 & 91.44 & 91.04 \\
\hline
\multicolumn{5}{l}{\textbf{MedXpertQA Text}}\\
Reasoning & \textbf{56.96} (\textcolor{blue}{$\uparrow$26.33\%}) & 45.94 & 36.38 & 30.63 \\
Understanding & \textbf{54.84} (\textcolor{blue}{$\uparrow$25.30\%}) & 43.80 & 33.96 & 29.54 \\
\hline
\multicolumn{5}{l}{\textbf{MMLU}}\\
Anatomy & \textbf{92.59} (\textcolor{blue}{$\uparrow$1.48\%}) & 92.59 & 88.15 & 91.11 \\
Clinical Knowledge & \textbf{95.09} (\textcolor{blue}{$\uparrow$2.64\%}) & 91.32 & 89.81 & 92.45 \\
College Biology & \textbf{99.31} (\textcolor{blue}{$\uparrow$2.09\%}) & 99.31 & 97.92 & 97.22 \\
College Medicine & \textbf{91.91} (\textcolor{blue}{$\uparrow$1.74\%}) & 88.44 & 85.55 & 90.17 \\
Medical Genetics & \textbf{100.00} (\textcolor{blue}{$\uparrow$4.00\%}) & 99.00 & 98.00 & 96.00 \\
Professional Medicine & \textbf{97.79} (\textcolor{blue}{$\uparrow$1.10\%}) & 97.43 & 96.69 & 96.69 \\
\hline
\end{tabular}}
\end{table}

\subsection{Performance of GPT-5 on USMLE Self Assessment}

As shown in Table \ref{tab:usmle-sample}, GPT-5 outperformed all baselines on all three steps, with the largest margin on Step 2 (+4.17\%). Step 2 focuses on clinical decision-making and management, aligning with GPT-5’s improved CoT reasoning. The average score across steps reached 95.22\% (+2.88\% vs GPT-4o), exceeding typical human passing thresholds by a wide margin, demonstrating the model’s readiness for high-stakes clinical reasoning tasks.

\begin{table}[htbp]
\centering
\caption{USMLE Sample Exam Performance (\%). The blue numbers and arrows indicate changes compared to GPT-4o-2024-11-20.}
\label{tab:usmle-sample}
\adjustbox{width=\textwidth,center}{
\begin{tabular}{lcccc}
\hline
 & GPT-5 & GPT-5-mini & GPT-5-nano & GPT-4o-2024-11-20 \\
\hline
Step 1 & \textbf{93.28} (\textcolor{blue}{$\uparrow$0.84\%}) & 93.28 & 93.28 & 92.44 \\
Step 2 & \textbf{97.50} (\textcolor{blue}{$\uparrow$4.17\%}) & 95.83 & 90.00 & 93.33 \\
Step 3 & \textbf{94.89} (\textcolor{blue}{$\uparrow$3.65\%}) & 94.89 & 92.70 & 91.24 \\
\hline
Average & \textbf{95.22} (\textcolor{blue}{$\uparrow$2.88\%}) & 94.67 & 91.99 & 92.34 \\
\hline
\end{tabular}}
\end{table}

\subsection{Performance of GPT-5 on VQA Benchmarks}

For multimodal reasoning (Table \ref{tab:mm-rad-results}), GPT-5 achieved a dramatic leap in MedXpertQA MM, with reasoning and understanding gains of +29.26\% and +26.18\%, respectively, relative to GPT-4o. This magnitude of improvement suggests significantly enhanced integration of visual and textual cues. 

However, in VQA-RAD, GPT-5 scored 70.92\%, slightly below GPT-5-mini (74.90\%). Given VQA-RAD’s relatively small scale and radiology-specific nature, this difference may reflect dataset-specific overfitting in the smaller model or conservative reasoning in GPT-5. A representative example from the MedXpertQA MM benchmark (Figure~\ref{fig:MM-1993-answer}) illustrates GPT-5’s capability to synthesize multimodal information in a clinically coherent manner. 

In this case, the model correctly identified esophageal perforation (Boerhaave syndrome) as the most likely diagnosis based on the combination of CT imaging findings, laboratory values, and key physical signs (suprasternal crepitus, blood-streaked emesis) following repeated vomiting. 
It then recommended a Gastrografin swallow study as the next management step, while explicitly ruling out other options and justifying each exclusion. This output demonstrates the model’s ability to integrate visual evidence with complex narrative context, maintain a structured diagnostic reasoning chain, and arrive at a high-stakes clinical decision that aligns with expert consensus.

\begin{table}[htbp]
\centering
\caption{Performance on VQA benchmarks (\%). The blue numbers and arrows indicate changes compared to GPT-4o-2024-11-20.}
\label{tab:mm-rad-results}
\adjustbox{width=\textwidth,center}{
\begin{tabular}{lcccc}
\hline
Dataset & GPT-5 & GPT-5-mini & GPT-5-nano & GPT-4o-2024-11-20 \\
\hline
\multicolumn{5}{l}{\textbf{MedXpertQA MM}}\\
Reasoning     & \textbf{69.99} (\textcolor{blue}{$\uparrow$29.26\%}) & 60.51 & 45.44 & 40.73 \\
Understanding & \textbf{74.37} (\textcolor{blue}{$\uparrow$26.18\%}) & 61.37 & 45.85 & 48.19 \\
\hline
\multicolumn{5}{l}{\textbf{Radiology}}\\
VQA-RAD & 70.92 & \textbf{74.90} (\textcolor{blue}{$\uparrow$4.99\%}) & 65.34 & 69.91 \\
\hline
\end{tabular}}
\end{table}

\begin{figure}[htbp]
\centering
\begin{tcolorbox}[
    colback=promptgray,
    colframe=black!50!black,
    colbacktitle=white!50!black,
    title=GPT-5 Reasoning and Answer for MedXpertQA: Case MM-1993,
    fontupper=\normalsize,
    enhanced,
    overlay={
        \node[anchor=north east, inner sep=3pt, yshift=-15pt] 
        at (frame.north east) {\includegraphics[width=0.25\textwidth]{MM-1993-a.jpeg}};
    } 
]
{\linespread{0.95}\selectfont
\begin{verbatim}
{
    "role": "assistant",
    "content": "Answer: E) Gastrografin swallow

    Reasoning:
    - He likely has acute alcoholic pancreatitis (epigastric 
        pain, elevated lipase, CT changes), which led to re-
        peated vomiting.
    - New suprasternal crepitus with blood-streaked emesis 
        after forceful vomiting strongly suggests esophageal 
        perforation (Boerhaave syndrome), not just a Mallory-
        Weiss tear.
    - The next step to confirm suspected perforation is a water-soluble contrast 
        esophagram (Gastrografin swallow). If positive, proceed with NPO, broad-
        spectrum antibiotics, and urgent surgical consultation.
    
    Why not the others:
    - A) Ondansetron: antiemetic alone misses a life-threatening perforation.
    - B) Folate/B12: macrocytosis can be addressed later; not the acute issue.
    - C) Supportive therapy: appropriate for pancreatitis, but new crepitus 
        requires urgent evaluation for perforation.
    - D) Epinephrine: for anaphylaxis or endoscopic hemostasis, not indicated 
        here."
},
{
    "role": "assistant",
    "content": "E) Gastrografin swallow"
}
\end{verbatim}
}
\end{tcolorbox}
\caption{GPT-5 reasoning output and final answer for MedXpertQA: case MM-1993.}
\label{fig:MM-1993-answer}
\end{figure}

\subsection{Comparison with human experts}

Table~\ref{tab:medxpertqa-text-mm} further shows a striking contrast in performance between
GPT-4o-2024-11-20, pre-licensed human experts, and GPT-5. GPT-4o performed below pre-licensed human experts on most dimensions, underperforming
by 5.03--15.90\% across reasoning and understanding in both text and multimodal settings.
In sharp contrast, GPT-5 not only closes this gap but surpasses human experts by a
substantial margin, achieving improvements of +15.22\% (text reasoning), +9.40\% (text
understanding), +24.23\% (multimodal reasoning), and +29.40\% (multimodal understanding).
These improvements are substantial, marking a notable advancement in model capability, shifting GPT-5 from human-comparable performance to consistently exceeding that of trained medical professionals in standardized benchmark evaluations.

The magnitude of this lead is particularly striking in multimodal settings, where GPT-5’s unified vision-language reasoning pipeline appears to deliver an integration of textual and visual evidence that even experienced clinicians struggle to match under time-limited test
conditions. This marked improvement from GPT-4o's below-human results to GPT-5's above-human performance highlights a significant advancement in LLM capabilities, with important potential implications for their use in real-world clinical decision support.

\begin{table}[htbp]
\centering
\caption{Comparison with human experts (Text and Multimodal)}
\label{tab:medxpertqa-text-mm}
\begin{tabular}{lcccccc}
\hline
Model & \multicolumn{3}{c}{MedXpertQA Text} & \multicolumn{3}{c}{MedXpertQA MM} \\
\hline
 & Reasoning & Understanding & Avg & Reasoning & Understanding & Avg \\
\hline
Expert (Pre-Licensed) & 41.74 & 45.44 & 42.60 & 45.76 & 44.97 & 45.53 \\
\hline
GPT-4o-2024-11-20   
& \makecell{30.63\\ (\textcolor{red}{$\downarrow$11.11\%})}
& \makecell{29.54\\ (\textcolor{red}{$\downarrow$15.90\%})}
& \makecell{30.37\\ (\textcolor{red}{$\downarrow$12.23\%})}
& \makecell{40.73\\ (\textcolor{red}{$\downarrow$5.03\%})}
& \makecell{48.19\\ (\textcolor{blue}{$\uparrow$3.22\%})}
& \makecell{42.80\\ (\textcolor{red}{$\downarrow$2.73\%})} \\
GPT-5-nano
& \makecell{36.38\\ (\textcolor{red}{$\downarrow$5.36\%})}
& \makecell{33.96\\ (\textcolor{red}{$\downarrow$11.48\%})}
& \makecell{35.17\\ (\textcolor{red}{$\downarrow$7.43\%})}
& \makecell{45.44\\ (\textcolor{red}{$\downarrow$0.32\%})}
& \makecell{45.85\\ (\textcolor{blue}{$\uparrow$0.88\%})}
& \makecell{45.65\\ (\textcolor{blue}{$\uparrow$0.12\%})} \\
GPT-5-mini
& \makecell{45.94\\ (\textcolor{blue}{$\uparrow$4.20\%})}
& \makecell{43.80\\ (\textcolor{red}{$\downarrow$1.64\%})}
& \makecell{44.87\\ (\textcolor{blue}{$\uparrow$2.27\%})}
& \makecell{60.51\\ (\textcolor{blue}{$\uparrow$14.75\%})}
& \makecell{61.37\\ (\textcolor{blue}{$\uparrow$16.40\%})}
& \makecell{60.94\\ (\textcolor{blue}{$\uparrow$15.41\%})} \\
\hline
GPT-5                 
& \makecell{\textbf{56.96} \\ (\textcolor{blue}{$\uparrow$15.22\%})} 
& \makecell{\textbf{54.84} \\ (\textcolor{blue}{$\uparrow$9.40\%})} 
& \makecell{\textbf{55.90} \\ (\textcolor{blue}{$\uparrow$13.30\%})} 
& \makecell{\textbf{69.99} \\ (\textcolor{blue}{$\uparrow$24.23\%})} 
& \makecell{\textbf{74.37} \\ (\textcolor{blue}{$\uparrow$29.40\%})} 
& \makecell{\textbf{72.18} \\ (\textcolor{blue}{$\uparrow$26.65\%})} \\

\hline
\end{tabular}
\end{table}

\section{Disscusion}
We evaluate the reasoning capabilities of the GPT-5 family of models on a wide range of multimodal tasks, revealing several key findings:

First, GPT-5 delivers substantial gains in multimodal medical reasoning, especially in datasets like MedXpertQA MM that demand tight integration of image-derived evidence with textual patient data. The observed improvements of +26–36\% over GPT-4o in multimodal settings suggest enhancements in cross-modal attention and alignment within the model's architecture or training.

Second, these gains are most pronounced in reasoning-intensive tasks, as evidenced by results from MedXpertQA Text and USMLE Step 2. Here, chain-of-thought (CoT) prompting likely synergizes with GPT-5’s enhanced internal reasoning capacity, enabling more accurate multi-hop inference. In contrast, in domains with high baseline accuracy (e.g., MMLU factual subtests), we note smaller but consistent improvements,  indicating that GPT-5's primary strength lies in its ability to tackle complex reasoning challenges rather than simply recalling facts.

Third, performance relative to humans is particularly noteworthy. GPT-5 not only matches but surpasses the performance of pre-licensed medical professionals in controlled QA/VQA evaluations, which raises both potential benefits and caution. On one hand, it underscores the potential for LLMs to serve as clinical decision-support systems; on the other hand, it is important to recognize that these evaluations occur within idealized, standardized testing environments that do not fully encompass the complexity, uncertainty, and ethical considerations inherent in real-world medical practice.

An unexpected observation is that GPT-5 scored slightly lower on VQA-RAD compared to its smaller counterpart, GPT-5-mini. This discrepancy may be attributed to scaling-related differences in reasoning calibration; larger models might adopt a more cautious approach in selecting answers for smaller datasets, resulting in fewer, albeit more conservative, correct predictions. Future research could explore adaptive prompting or calibration techniques specifically tailored for small-domain multimodal tasks.

\section{Conclusion}
This study presents the first controlled, longitudinal evaluation of GPT-5’s capabilities in multimodal medical reasoning, comparing its performance to GPT-4o-2024-11-20, smaller GPT-5 variants, and human experts under standardized zero-shot CoT prompting. Across diverse QA and VQA benchmarks, GPT-5 demonstrates substantial and consistent gains, particularly in reasoning-intensive and multimodal tasks. Notably, the model’s ability to surpass trained medical professionals on MedXpertQA MM by large margins signifies a qualitative shift in LLM capabilities, moving from near-human performance in GPT-4o-2024-11-20 to clear super-human proficiency. These results highlight GPT-5’s potential as a reliable core component for multimodal clinical decision support, capable of integrating complex textual and visual information streams to produce accurate, well-justified recommendations. However, it is important to note that the benchmarks used reflect idealized testing conditions and may not fully capture the variability, uncertainty, and ethical considerations of real-world practice. Future work should investigate prospective clinical trials, domain-adapted fine-tuning strategies, and calibration methods to ensure safe and transparent deployment. Ultimately, the advancements represented by GPT-5 mark a pivotal moment in the evolution of medical AI, bridging the gap between research prototypes and practical, high-impact clinical tools.

\bibliographystyle{unsrt}
\bibliography{newbib}

\begin{thebibliography}{10}

\bibitem{achiam2023gpt}
Josh Achiam, Steven Adler, Sandhini Agarwal, Lama Ahmad, Ilge Akkaya,
  Florencia~Leoni Aleman, Diogo Almeida, Janko Altenschmidt, Sam Altman,
  Shyamal Anadkat, et~al.
\newblock Gpt-4 technical report.
\newblock {\em arXiv preprint arXiv:2303.08774}, 2023.

\bibitem{liu2024deepseek}
Aixin Liu, Bei Feng, Bing Xue, Bingxuan Wang, Bochao Wu, Chengda Lu, Chenggang
  Zhao, Chengqi Deng, Chenyu Zhang, Chong Ruan, et~al.
\newblock Deepseek-v3 technical report.
\newblock {\em arXiv preprint arXiv:2412.19437}, 2024.

\bibitem{openai2025gpt5}
OpenAI.
\newblock Introducing gpt-5, August 7 2025.

\bibitem{azar2025llm}
William~S Azar, Dylan~M Junkin, Charles Hesswani, Christopher~R Koller, Sahil~H
  Parikh, Kyle~C Schuppe, Nicholas Williams, Daniel Nethala, Neil Mendhiratta,
  Alexander~P Kenigsberg, et~al.
\newblock Llm-mediated data extraction from patient records after radical
  prostatectomy.
\newblock {\em NEJM AI}, 2(6):AIcs2400943, 2025.

\bibitem{singhal2023large}
Karan Singhal, Shekoofeh Azizi, Tao Tu, S~Sara Mahdavi, Jason Wei, Hyung~Won
  Chung, Nathan Scales, Ajay Tanwani, Heather Cole-Lewis, Stephen Pfohl, et~al.
\newblock Large language models encode clinical knowledge.
\newblock {\em Nature}, 620(7972):172--180, 2023.

\bibitem{guo2023gpt4graph}
Jiayan Guo, Lun Du, Hengyu Liu, Mengyu Zhou, Xinyi He, and Shi Han.
\newblock Gpt4graph: Can large language models understand graph structured
  data? an empirical evaluation and benchmarking.
\newblock {\em arXiv preprint arXiv:2305.15066}, 2023.

\bibitem{wang2025triad}
Shansong Wang, Mojtaba Safari, Qiang Li, Chih-Wei Chang, Richard~LJ Qiu, Justin
  Roper, David~S Yu, and Xiaofeng Yang.
\newblock Triad: Vision foundation model for 3d magnetic resonance imaging.
\newblock {\em arXiv preprint arXiv:2502.14064}, 2025.

\bibitem{wang2025unifying}
Shansong Wang, Zhecheng Jin, Mingzhe Hu, Mojtaba Safari, Feng Zhao, Chih-Wei
  Chang, Richard~LJ Qiu, Justin Roper, David~S Yu, and Xiaofeng Yang.
\newblock Unifying biomedical vision-language expertise: Towards a generalist
  foundation model via multi-clip knowledge distillation.
\newblock {\em arXiv preprint arXiv:2506.22567}, 2025.

\bibitem{liu2025application}
Fenglin Liu, Hongjian Zhou, Boyang Gu, Xinyu Zou, Jinfa Huang, Jinge Wu, Yiru
  Li, Sam~S Chen, Yining Hua, Peilin Zhou, et~al.
\newblock Application of large language models in medicine.
\newblock {\em Nature Reviews Bioengineering}, pages 1--20, 2025.

\bibitem{nori2023capabilities}
Harsha Nori, Nicholas King, Scott~Mayer McKinney, Dean Carignan, and Eric
  Horvitz.
\newblock Capabilities of gpt-4 on medical challenge problems.
\newblock {\em arXiv preprint arXiv:2303.13375}, 2023.

\bibitem{thirunavukarasu2023large}
Arun~James Thirunavukarasu, Darren Shu~Jeng Ting, Kabilan Elangovan, Laura
  Gutierrez, Ting~Fang Tan, and Daniel Shu~Wei Ting.
\newblock Large language models in medicine.
\newblock {\em Nature medicine}, 29(8):1930--1940, 2023.

\bibitem{hu2024advancing}
Mingzhe Hu, Joshua Qian, Shaoyan Pan, Yuheng Li, Richard~LJ Qiu, and Xiaofeng
  Yang.
\newblock Advancing medical imaging with language models: featuring a spotlight
  on chatgpt.
\newblock {\em Physics in Medicine \& Biology}, 69(10):10TR01, 2024.

\bibitem{openai2023gpt35}
OpenAI.
\newblock Introducing gpt-3.5, March 1 2023.

\bibitem{yang2023performance}
Zhichao Yang, Zonghai Yao, Mahbuba Tasmin, Parth Vashisht, Won~Seok Jang,
  Feiyun Ouyang, Beining Wang, Dan Berlowitz, and Hong Yu.
\newblock Performance of multimodal gpt-4v on usmle with image: potential for
  imaging diagnostic support with explanations.
\newblock {\em medRxiv}, pages 2023--10, 2023.

\bibitem{hopkins2023chatgpt}
Benjamin~S Hopkins, Vincent~N Nguyen, Jonathan Dallas, Pavlos Texakalidis, Max
  Yang, Alex Renn, Gage Guerra, Zain Kashif, Stephanie Cheok, Gabriel Zada,
  et~al.
\newblock Chatgpt versus the neurosurgical written boards: a comparative
  analysis of artificial intelligence/machine learning performance on
  neurosurgical board--style questions.
\newblock {\em Journal of Neurosurgery}, 139(3):904--911, 2023.

\bibitem{yeo2023assessing}
Yee~Hui Yeo, Jamil~S Samaan, Wee~Han Ng, Peng-Sheng Ting, Hirsh Trivedi, Aarshi
  Vipani, Walid Ayoub, Ju~Dong Yang, Omer Liran, Brennan Spiegel, et~al.
\newblock Assessing the performance of chatgpt in answering questions regarding
  cirrhosis and hepatocellular carcinoma.
\newblock {\em Clinical and molecular hepatology}, 29(3):721, 2023.

\bibitem{johnson2023assessing}
Douglas Johnson, Rachel Goodman, J~Patrinely, Cosby Stone, Eli Zimmerman,
  Rebecca Donald, Sam Chang, Sean Berkowitz, Avni Finn, Eiman Jahangir, et~al.
\newblock Assessing the accuracy and reliability of ai-generated medical
  responses: an evaluation of the chat-gpt model.
\newblock {\em Research square}, pages rs--3, 2023.

\bibitem{bhayana2023gpt}
Rajesh Bhayana, Robert~R Bleakney, and Satheesh Krishna.
\newblock Gpt-4 in radiology: improvements in advanced reasoning.
\newblock {\em Radiology}, 307(5):e230987, 2023.

\bibitem{omar2024chatgpt}
Mohamed Omar, Varun Ullanat, Massimo Loda, Luigi Marchionni, and Renato Umeton.
\newblock Chatgpt for digital pathology research.
\newblock {\em The Lancet Digital Health}, 6(8):e595--e600, 2024.

\bibitem{demir2024enhancing}
Gizem~Bozta{\c{s}} Demir, Ya{\u{g}}{\i}zalp S{\"u}k{\"u}t, G{\"o}khan~Serhat
  Duran, K{\"u}bra~G{\"u}lnur Topsakal, and Serkan G{\"o}rg{\"u}l{\"u}.
\newblock Enhancing systematic reviews in orthodontics: a comparative
  examination of gpt-3.5 and gpt-4 for generating pico-based queries with
  tailored prompts and configurations.
\newblock {\em European Journal of Orthodontics}, 46(2):cjae011, 2024.

\bibitem{tung2024comparison}
Joshua Yi~Min Tung, Sunil~Ravinder Gill, Gerald Gui~Ren Sng, Daniel Yan~Zheng
  Lim, Yuhe Ke, Ting~Fang Tan, Liyuan Jin, Kabilan Elangovan, Jasmine
  Chiat~Ling Ong, Hairil~Rizal Abdullah, et~al.
\newblock Comparison of the quality of discharge letters written by large
  language models and junior clinicians: single-blinded study.
\newblock {\em Journal of medical Internet research}, 26:e57721, 2024.

\bibitem{kweon2024ehrnoteqa}
Sunjun Kweon, Jiyoun Kim, Heeyoung Kwak, Dongchul Cha, Hangyul Yoon, Kwang Kim,
  Jeewon Yang, Seunghyun Won, and Edward Choi.
\newblock Ehrnoteqa: An llm benchmark for real-world clinical practice using
  discharge summaries.
\newblock {\em Advances in Neural Information Processing Systems},
  37:124575--124611, 2024.

\bibitem{hao2025large}
Yuexing Hao, Zhiwen Qiu, Jason Holmes, Corinna~E L{\"o}ckenhoff, Wei Liu,
  Marzyeh Ghassemi, and Saleh Kalantari.
\newblock Large language model integrations in cancer decision-making: a
  systematic review and meta-analysis.
\newblock {\em npj Digital Medicine}, 8(1):450, 2025.

\bibitem{jin2020disease}
Di~Jin, Eileen Pan, Nassim Oufattole, Wei-Hung Weng, Hanyi Fang, and Peter
  Szolovits.
\newblock What disease does this patient have? a large-scale open domain
  question answering dataset from medical exams.
\newblock {\em arXiv preprint arXiv:2009.13081}, 2020.

\bibitem{hendryckstest2021}
Dan Hendrycks, Collin Burns, Steven Basart, Andy Zou, Mantas Mazeika, Dawn
  Song, and Jacob Steinhardt.
\newblock Measuring massive multitask language understanding.
\newblock {\em Proceedings of the International Conference on Learning
  Representations (ICLR)}, 2021.

\bibitem{lau2018dataset}
Jason~J Lau, Soumya Gayen, Asma Ben~Abacha, and Dina Demner-Fushman.
\newblock A dataset of clinically generated visual questions and answers about
  radiology images.
\newblock {\em Scientific data}, 5(1):1--10, 2018.

\bibitem{zuo2025medxpertqa}
Yuxin Zuo, Shang Qu, Yifei Li, Zhangren Chen, Xuekai Zhu, Ermo Hua, Kaiyan
  Zhang, Ning Ding, and Bowen Zhou.
\newblock Medxpertqa: Benchmarking expert-level medical reasoning and
  understanding.
\newblock {\em arXiv preprint arXiv:2501.18362}, 2025.

\bibitem{kung2023performance}
Tiffany~H Kung, Morgan Cheatham, Arielle Medenilla, Czarina Sillos, Lorie
  De~Leon, Camille Elepa{\~n}o, Maria Madriaga, Rimel Aggabao, Giezel
  Diaz-Candido, James Maningo, et~al.
\newblock Performance of chatgpt on usmle: potential for ai-assisted medical
  education using large language models.
\newblock {\em PLoS digital health}, 2(2):e0000198, 2023.

\end{thebibliography}

\end{document}